\documentclass[12pt,a4paper]{article}

\usepackage[british]{babel}

\usepackage[a4paper,top=2cm,bottom=2cm,left=2.5cm,right=2.5cm,marginparwidth=1.75cm]{geometry}

\usepackage[style=authoryear,sorting=nyt,backend=biber]{biblatex}

\DeclareLanguageMapping{british}{british-apa} 

\usepackage{amsmath}
\usepackage{graphicx}
\usepackage[colorlinks=true, allcolors=blue]{hyperref}
\usepackage{hyperref}
\usepackage[title]{appendix}
  
\usepackage{mathrsfs}
\usepackage{amsfonts}
\usepackage{booktabs} 
\usepackage{caption}  
\usepackage{threeparttable} 
\usepackage{algorithm}
\usepackage{algorithmicx}
\usepackage{algpseudocode}
\usepackage{listings}
\usepackage{enumitem}
\usepackage{chngcntr}
\usepackage{booktabs}
\usepackage{lipsum}
\usepackage{subcaption}
\usepackage{authblk}
\usepackage[T1]{fontenc}    
\usepackage{csquotes}       
\usepackage{diagbox}
\usepackage{comment}
\usepackage{enumitem}

\usepackage{setspace}
\usepackage{titlesec}
\titleformat{\section} 
  {\normalfont\Large\bfseries}{\thesection.}{1em}{}
  
\usepackage{float}   
\usepackage{caption} 


\title{Operational evaluation of data-driven forest fire forecasting models}

\author[1*]{Shahbaz Alvi} 
\author[1]{Jose Maria Costa Saura} 
\author[1,2]{Italo Epicoco} 

\affil[1]{\small CMCC Foundation - Euro-Mediterranean Center on Climate Change, Italy}
\affil[2]{\small Department of Engineering for Innovation, University of Salento, Italy}

\affil[*]{\small Corresponding author: \texttt{shahbaz.alvi@cmcc.it}}

\date{}  

\newcommand{\addref}[1]{\textcolor{red}{[Add reference here]}}

\begin{document}
\maketitle

\begin{abstract}
Within the standard paradigm, machine learning models for daily wildfire forecasting are typically evaluated using standard metrics such as Recall, Precision, and F1 computed on a curated, dedicated test dataset. We argue that this is structurally inadequate for operational assessment and, additionally, less transparent for non-experts such as firefighters and decision-makers. In particular, test dataset Recall remains an approximate proxy for the accuracy of operational fire activity forecast, whereas test dataset Precision offers little operationally relevant insight concerning the region-wide false-alarm distribution. In this manuscript, we propose a novel evaluation framework in which daily fire activity forecast maps — previously used primarily for qualitative illustration over a handful of days — become the central component of model evaluation, mimicking a wildfire management workflow. We evaluate the daily recall - the fraction of actual fire pixels correctly flagged on a given day - for every day in the test period with at least one fire event, and assesses the region-wide false-alarm behavior across all days in the test period without recorded fires. We apply this evaluation framework to a Basic CNN, two deeper CNN variants, and a ConvLSTM trained on data from Greece and the surrounding region. In the test dataset-based evaluation, the ConvLSTM architecture is ranked the lowest, while in our framework, it achieves near-perfect detection (daily recall $> 0.99$ on 40\% of the days) and produces the lowest false alarm rate across the entire region, highlighting that the test dataset-based evaluation can misrank the operational performance of the model for field deployment. We further show that ensemble averaging reduces the region-wide false-alarm rate.

\end{abstract}




\section{Introduction}

Wildfires are becoming an increasing concern under climate change, as rising temperatures, prolonged droughts, and more frequent heat extremes intensify fire weather and extend fire seasons across many regions \parencite{OECD2023Wildfires}. This escalation in fire activity places growing pressure on operational wildfire management, demanding more accurate fire activity forecasting for the Early Warning Systems (EWS) to mitigate the risks and reduce impacts on lives, infrastructure, and essential services.  For many decades, process-based methods, based on meteorological data, have been used to assess the wildfire danger.

The recent surge in artificial intelligence and machine learning research has accelerated the development in the field of wildfire management - see \parencite{p.jain.2020} for a review. Machine learning approaches demonstrate improved accuracy and provide more localized forest fire forecasts, thus reducing false-positive rates, compared to process-based canonical methods, such as the Canadian Fire Weather Index (FWI) \parencite {DiGiuseppe2025,Kondylatos2022}. Several machine learning architectures have been applied to the problem of forecasting the next-day fire activity, such as Random Forest (RF) ~\parencite{Stojanova2012,cao2017,yu2017}, Artificial Neural Networks (ANN) ~\parencite{VGarcia1996,Alonso2002,dutta2013}, and deep learning architectures \parencite{ji2024global,mambile2024deep,pais2021deep,Kondylatos2022,DiGiuseppe2025,ramayanti2024wildfire,he2024deep,zakari2025spatio,deng2023wildfire,jiang2024wildfire}. Across these studies, model evaluation has relied on the standard paradigm where evaluation is based on metrics such as Recall, Precision, or ROC evaluated on a dedicated test-dataset, while region-wdie model inference has primarily been used for illustration purposes.

Daily fire activity maps highlighting regions of high fire likelihood are fundamental to the operational workflow of forest fire management adopted by the civil protection agencies and forest management authorities. Recognizing the importance of the fire activity maps in the operational workflow of forest fire management, we build on the work of \parencite{Kondylatos2022} and further propose a model evaluation framework based on region-wide inference. In particular, we argue that the model evaluation on a curated test dataset reflects the model's performance exclusively in the context of the test dataset. While essential in monitoring model training, the test dataset Recall might not represent the real-world large-scale operational accuracy of the model, while the Precision gives a limited picture of the model's region-wide false alarm behavior. Standard evaluation metrics from a test dataset are also less transparent and less interpretable for the civil protection agencies. Identifying the gap between the standard model evaluation framework and the operational fire management workflow, our proposed framework puts the fire activity forecast map at the center of the evaluation process, where not only the model's forecast accuracy in predicting fires is evaluated, but also the model's operational false-alarm behavior becomes directly observable. These aspects of model performance are crucial before ML models are integrated into the existing firefighting operational chain.

This manuscript is structured in the following way: In Section \ref{sec:problem_intro}, we explain the architecture of the neural networks considered in this work, the dataset and the target geographic region, and our sampling strategy and training methodology. In Section \ref{sec:results}, we present the results of our analysis. In Sections \ref{sec:discussion} and \ref{sec:conclusions}, we discuss our findings and present our conclusions, respectively.

\section{Methodology}\label{sec:problem_intro}
The CNN \parencite{Krizhevsky2012AlexNet} and ConvLSTM \parencite{shi2015convolutional} architectures have been explored for forecasting fire activity at global (e.g. \parencite{ji2024global}) and regional scale (e.g. \parencite{Kondylatos2022}). CNNs can capture complex nonlinear relationships in heterogeneous data, such as the interplay between topography, vegetation, and human activities for fire occurrence, but a vanilla CNN does not account for temporal correlation. Several studies report improved performance with a ConvLSTM, an amalgamation of CNN and LSTM architectures \parencite{Kondylatos2022,mambile2024deep,ji2024global}. The hybrid architecture of ConvLSTM efficiently extracts spatiotemporal correlations, which are important for forecasting fire occurrence as fire predictors vary on different time scales, e.g., weather on a daily scale, vegetation typically on a 10-day scale, and constant topographic variables.

The proposed method frames fire activity detection as a patch-based binary classification problem. The input image is partitioned into overlapping tiles, each centered on a target pixel that is assigned a binary label (fire / no-fire). The networks are trained on this labeled tile dataset. At inference time, the trained network is applied to all tiles extracted from a test image, and the resulting class probabilities are mapped back to their corresponding center pixels, yielding a dense probability map in which the likelihood of fire is encoded in each pixel.

\subsection{Architecture}\label{subsec:architecture}
In this work, we compared the performance of three variations of the vanilla CNN architecture and a ConvLSTM. The architecture and the model training were set up in the PyTorch Lightning framework \parencite{pytorch_lightning}, and MLFlow was utilized for tracking model training and logging. We trained the model with Adam optimizer \parencite{adam_optimizer} and the PyTorch ReduceLROnPlateau learning-rate scheduler. The output layer of the model is a \texttt{LogSoftmax} that returns the log of the normalized probabilities corresponding to the sample's assignment to the ``fire'' and ``no-fire'' labels. We used the negative log-likelihood as the loss function.

\begin{figure}[ht]
  \centering
  \includegraphics[width=\linewidth]{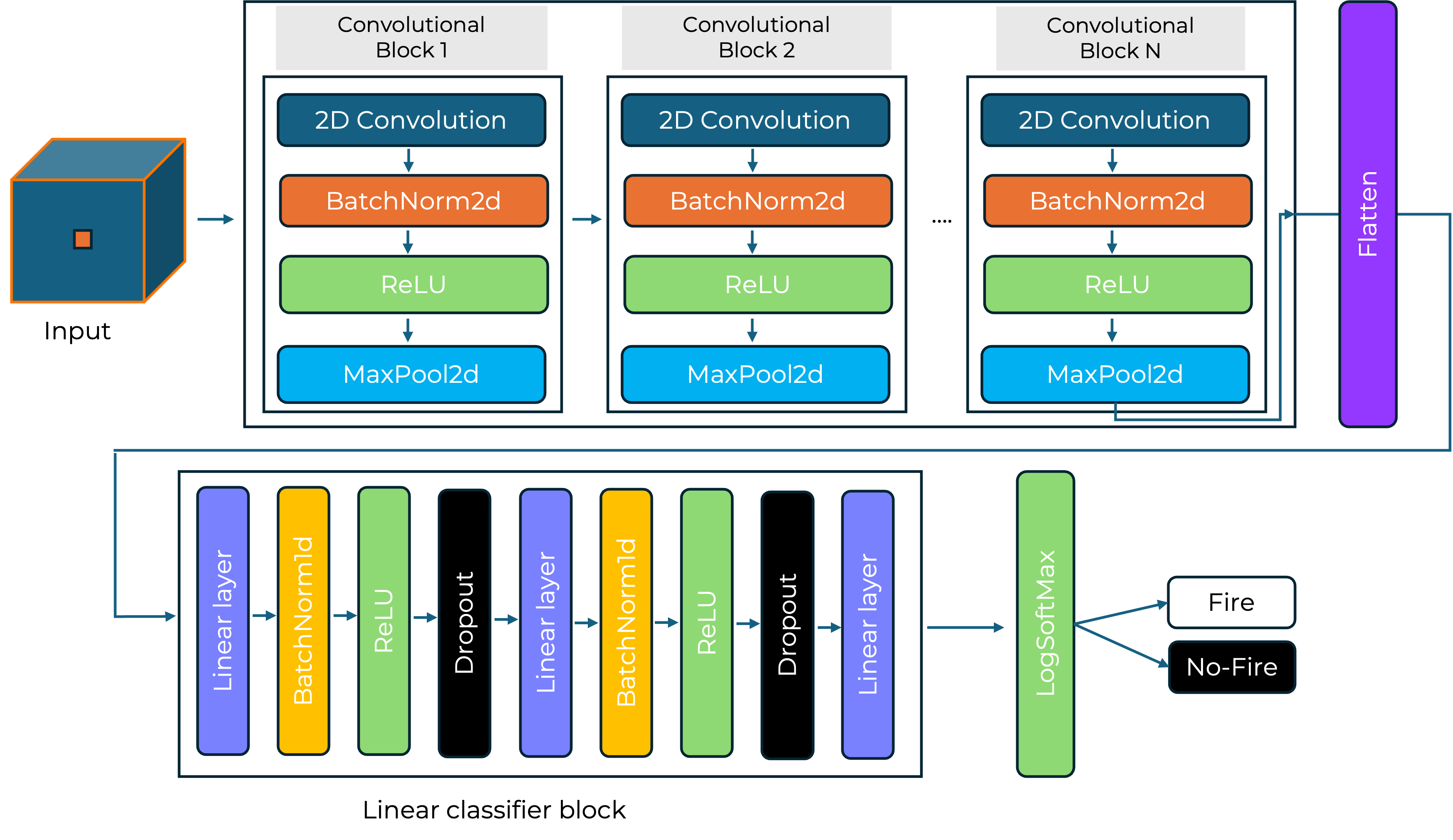}
  \caption{Architecture of the CNN considered in this manuscript. The deeper variations of the vanilla CNN differ in having more Convolutional blocks in the network.}
  \label{fig:cnn-network}
\end{figure}

\begin{figure}[ht]
  \centering
  \includegraphics[width=\linewidth]{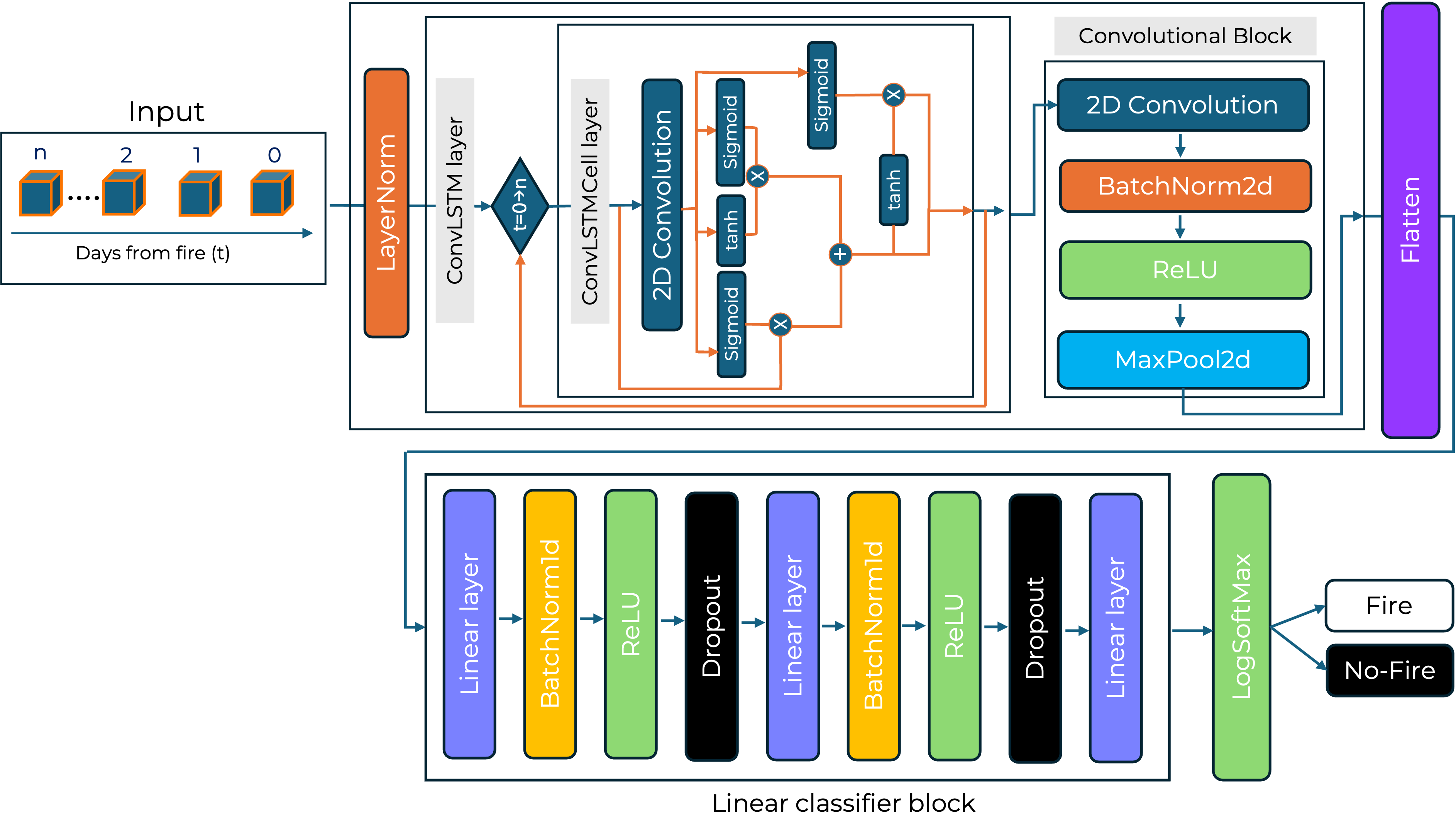}
  \caption{Architecture of the ConvLSTM considered in this manuscript. The architecture of the network is identical to that used and provided by \parencite{Kondylatos2022}.}
  \label{fig:convlstm-network}
\end{figure}

We have considered a vanilla CNN (\textit{Basic CNN}) - kernel size 3, and 18 hidden layer size - shown in Figure \ref{fig:cnn-network}, with 40.6K total number of trainable parameters. The Basic CNN is an efficient image classifier but lacks the depth to capture subtle, higher-order correlations across spatial features. Two deeper variants of the Basic CNN had thus been considered for comparison: A CNN with 3 convolution blocks and a single classifier (\textit{Deeper CNN 1}), consisting of a total of 66.5K trainable parameters, and one with 4 Convolution blocks and a classifier (\textit{Deeper CNN 2}), with a total of 111K trainable parameters. In this work, we use the ConvLSTM architecture as used in \parencite{Kondylatos2022} (see Figure \ref{fig:convlstm-network} for a schematic diagram) without modifications. The ConvLSTM consisted of a single ConvLSTMCell (which includes an LSTM block), followed by a convolution block and a classifier block. The total number of learnable parameters is 371K.
\subsection{Dataset and case study}\label{subsec:ds_study-area}
The training and validation samples are derived from the publicly available dataset published in \parencite{Kondylatos2022}. The high-resolution data cube \parencite{prapas2022firecube} is geographically centered on Greece and includes the Balkan peninsula and western Turkey, covering a total geographical footprint of 1,253~km $\times$ 983~km at a spatial resolution of 1~km and a daily temporal resolution. 

The dataset spans from 2009 to 2021 and includes 90 fire drivers and the target variable, the burn mask on a given day. The dataset also contains the Corine Land Cover (CLC) classes of the target region at 1 km resolution \parencite{clc_website}. The native 100 m CLC layer was aggregated to the 1 km grid by majority rule: each 1 km pixel is assigned the CLC class most frequent among the 100 m cells it contains. A ``remapped'' CLC is also provided in the dataset (for details on remapping the CLC classes and other details regarding the dataset, see the supplementary documents of \parencite{Kondylatos2022} and \parencite{prapas2022firecube}).

\subsection{Sampling strategy and training setup}\label{subsec:samples-training}
\subsubsection{Sampling strategy} \label{subsubsec:sampling}
Each sample patch is a multi-channel 25 $\times$ 25 ``image'' on which the classifier is trained. The model training approach adopted in this work is more operationally significant (in contrast, see XX), for two reasons. First, the model learns fire occurrence strictly with local-area information (e.g., vegetation, micro-climatic effects) within the image patch. Second, this approach conveniently allows building the training dataset with a balanced ratio of negative to positive samples. The sample preparation strategy is crucial in this approach as it not only affects model performance on the dedicated test dataset but also profoundly affects the inference maps of fire activity. However, a standard optimal recipe for sample curation from the large pool of available samples is unavailable, and various strategies have been adopted with some similarities (see \parencite{Kondylatos2022,ramayanti2024wildfire,he2024deep}). \parencite{Kondylatos2022} performed a Shapley analysis \parencite{lundberg2017shap} and selected best predictors for the model. Following their analysis, we retained all the variables used by \parencite{Kondylatos2022}, except for the Soil Moisture Index (SMI). While we acknowledge the importance of the SMI variable, we had excluded it due to lack of continuity in data availability, which would have impeded the operational deployment of the model. Therefore, in this work, we used 24 out of the 25 fire covariates used in \parencite{Kondylatos2022}, which are listed in Table \ref{tab:fire_preds}.
\begin{table}
    \caption{List of fire predictors used to forecast forest fire probability. More details on the remapping of original CLC classes can be found in \cite{Kondylatos2022}.}
    \centering
    \begin{tabular}{|c c c|}
        \hline
        \textbf{No.} & \textbf{Variable} & \textbf{Units} \\
        \hline
         1 & Normalized Difference Vegetation Index (NDVI) & -\\ 
        \hline
         2-3 & Day \& night Land Surface Temperature (LST) & K\\
        \hline
        4 & Daily maximum dew point temperature at 2-meters & K\\
        \hline
         5 & Daily maximum air temperature at 2-meters & K\\
        \hline
        6 & Daily maximum surface pressure at sea level & Pa\\
        \hline
        7  & Daily maximum total precipitation  & meters\\
        \hline
         8 & Daily maximum wind speed at 10-meters & m/s\\
        \hline
         9 & Minimum relative humidity & - \\
        \hline
         10 & Elevation (Digital Elevation Model - DEM) & m\\
        \hline
         11 & Slope & degrees\\
        \hline
         12 & Distance to nearest road & meters\\
        \hline
         13 & Distance to waterway & meters\\
        \hline
         14 & Population density & humans/Km$^2$\\
        \hline
         15 - 24 &  Remapped Corine Land Cover (CLC) & - \\
        \hline
    \end{tabular}

    \label{tab:fire_preds}
\end{table}

\begin{figure}[h]
  \centering
  \includegraphics[width=0.4\linewidth]{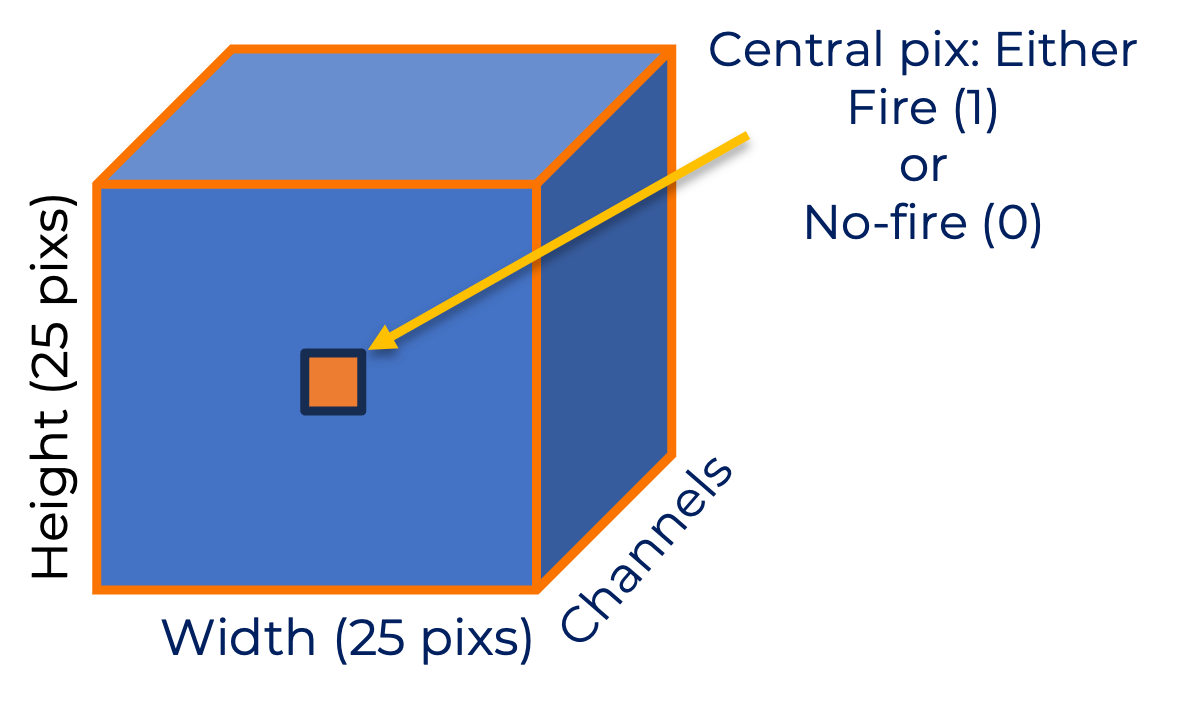}
  \vspace{0.5em}
  \includegraphics[width=0.5\linewidth]{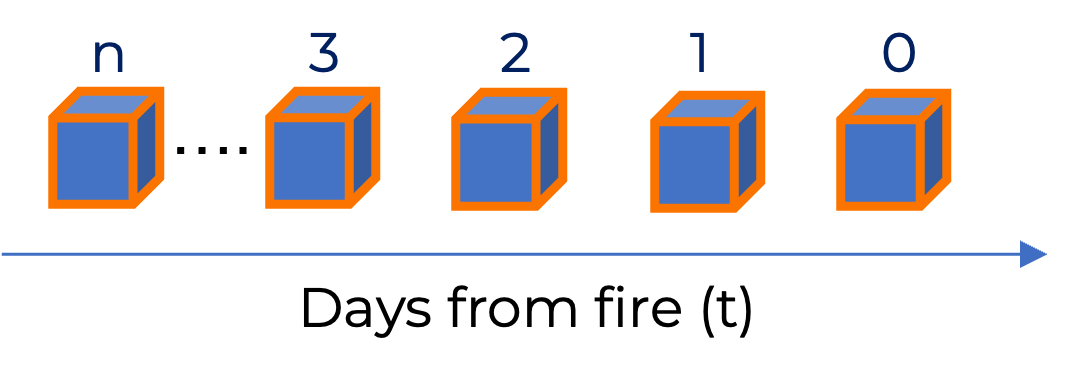}
  \caption{(Left) Figure shows the format of a single sample for the CNN network, which is a 3D tensor. (Right) A single sample of the ConvLSTM network also includes a time dimension of length 10, and it is a 4D tensor.}
  \label{fig:sample_format}
\end{figure}

For a CNN network, a single sample patch is a 3D tensor of shape 25 $\times$ 25 pixels with 24 channels (see Figure \ref{fig:sample_format}a). The samples for the ConvLSTM network have an additional time dimension; therefore, each sample is a 4D tensor (see Figure \ref{fig:sample_format}b). In the case of ConvLSTM, the temporal dimension of the sample spans the day of the event and 9 days before it (either fire or no-fire) - making the length of the temporal dimension equal to 10.

\begin{figure}
  \centering
  \includegraphics[width=1\linewidth]{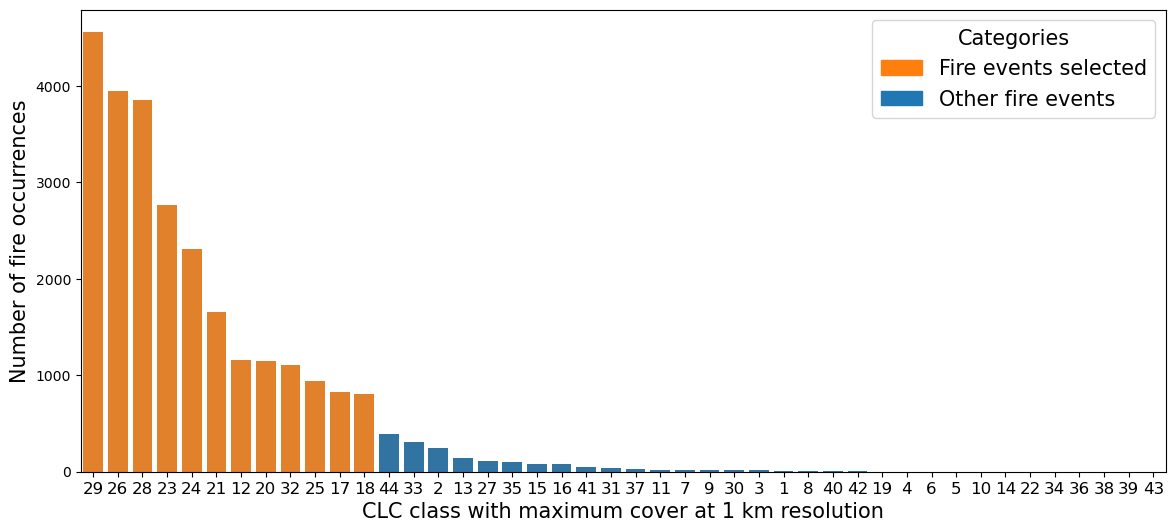}
  \caption{Plot showing the distribution of the fires in the entire dataset by their CLC class. On the x-axis are the CLC labels, and on the y-axis is the frequency of fire activity with that CLC class. The orange colored bars represent the distribution of the fire samples used for model training.}
  \label{fig:fire_clc_dist}
\end{figure}

Figure \ref{fig:fire_clc_dist} shows the frequency distribution of the fires in the entire dataset by the most frequent CLC class on 1 km resolution. We selected fire samples exclusively from CLC classes with higher fire occurrence frequency (shown in orange in Figure \ref{fig:fire_clc_dist}). This strategy was aimed at reducing the overestimation of false positives in region-wide inference, which can arise because the model had learned to associate fire activity with CLC classes where fires did not regularly occur. While fire occurrence in the Mediterranean region is concentrated in July, August, and September (henceforth referred to as JAS), fire samples are freely chosen from the entire year, having been restricted only to have a unique geographic location in a given year.
We identified 25,605 fire and over 250,000 no-fire samples. Fire occurrence is stochastic, and the conditions preceding past fires partially overlap those where no fires occurred. We had built the training dataset, balancing the similarities and differences between the two kinds of samples. No-fire samples that closely resemble fire samples systematically inflate predicted fire activity; we therefore curated the no-fire samples to control false positives on region-wide inference, using the following strategy:
\begin{enumerate}
    \item A 2:1 ratio is maintained between no-fire samples and fire samples in each year, balancing the number of positive and negative samples presented to the model during the training phase, while maintaining a sufficiently large distribution of no-fire samples in the feature space.
    \item Within each year, the no-fire samples were chosen from the period between the first and the last fire detection date (the fire season).
    \item Within the fire season, no-fire samples were drawn on the dates when no fires were recorded in the entire region of interest.   
    \item No-fire samples were drawn from locations where the dominant land cover type is theoretically susceptible to fires, even if no fires had been registered on that land cover in the dataset. This spans the land cover labels 12 to 32 (inclusive). 
    \item In a given year, all negative samples had unique locations, and they are mutually exclusive from the positive sample location.

\end{enumerate}
Following the outlined strategy, we drew  23,742 fire samples and 47,484 no-fire samples, which were split chronologically: samples from 2009 to 2018 were used for training, validation samples from 2019, and samples from 2020 and 2021 were reserved for testing. The model was trained for 150 epochs. The epoch-averaged validation loss was monitored during the training, and the network parameters corresponding to the lowest validation loss were retained as the best model. During the training, validation, and testing phases, we logged Recall, Precision, and the F1-score computed on the respective datasets. These scores are important for monitoring training progress and identifying potential bugs in the code implementation, and provide valuable insights into the model behavior and test the sampling strategy for dataset curation. 

\subsubsection{Training model ensemble}\label{subsubsec:model_ensemble_training}
Ensemble techniques, in which multiple base models are combined to outperform any single member of the ensemble, have a long history in numerical weather prediction (\parencite{Palmer1995, Weisheimer2009, DoblasReyes2020}) and in machine learning more broadly (\parencite{ensemble_review}). We train a seven-member ensemble of the ConvLSTM architecture, as it has been identified as the best-performing network (see Section \ref{sec:results}), each member initialized with independent random weights and trained on the identical dataset and procedure described previously for a single model.

\section{Results}\label{sec:results}
\subsection{Standard ML metrics}\label{subsec:standard_results}
In Table \ref{tab:test_scores}, the Recall, Precision, and F1-score computed on the test dataset using the best model configuration are shown. Table \ref{tab:test_scores} shows that F1-scores are comparable across all four architectures, reflecting a balance between Recall and Precision for the fire and no-fire classes, even though ConvLSTM has the lowest score among all architectures. Notably, increasing network depth — from Basic CNN to Deeper CNN 1 to Deeper CNN 2 — did not further improve the Recall.

\begin{table}
    \caption{The table reports the statistics computed on the samples from year 2020 and 2021, which were reserved for testing using the model configuration corresponding to the lowest validation loss.}
    \centering
    \begin{tabular}{|ccccc|}
         \hline
         \begin{tabular}{@{}c@{}}\textbf{Metric/}  \\ \textbf{Model} \end{tabular} & \textbf{Basic CNN} & \textbf{Deeper CNN 1 }& \textbf{Deeper CNN 2} & \textbf{ConvLSTM} \\
         \hline
         \textbf{Recall} & 0.89 & 0.90 & 0.90 & 0.87 \\
         \hline
         \textbf{Precision} & 0.77 & 0.77 & 0.78 & 0.73 \\
         \hline
         \textbf{F1-score} & 0.82 & 0.83 & 0.83 & 0.80 \\
         \hline
    \end{tabular}
    \label{tab:test_scores}
\end{table}

\subsection{Map-based model evaluation}\label{subsec:mapbased_eval}
Daily fire occurrence maps play a crucial role in operational wildfire management by enabling spatial differential analysis of fire occurrence likelihood and relative vulnerability. Therefore, systematic model performance on daily region-wide inference is crucial to ensure consistency between model evaluation and the decision-making framework in which the model is expected to be deployed. The performance metrics computed on the test dataset, while insightful during model training, are of limited use in daily operations, as their interpretation requires expertise in machine learning that firefighting units and civil protection personnel may not have. Beyond interpretability, the two evaluation settings are not equivalent. 
We introduce the metric \textit{daily recall} which is the per-pixel Recall computed on a single day: the fraction of fire pixels on a given day assigned fire probability > 0.5. Since it is computed over the entire fire mask on every day in the test period, daily recall is unaffected by the data curation of the test dataset. In particular, test dataset Recall and daily recall differ in coverage - the former is a proxy for the latter. Precision, however, depends on the negative class rate. The curated 2:1 negative-to-positive class ratio of the test set is far from the operational setting, where fire pixels are rare across the full map. Test-set Precision, therefore, reflects the performance on a curated dataset and says little about operational false-alarm behavior over the entire region, which motivates the map-based analysis explained in the following paragraph. 

Daily region-wide inference is computed over the fire season (JAS) in 2020 and 2021 (149 days total). The inference is restricted to locations with fire-susceptible land-cover classes; non-susceptible classes such as urban fabric, sea, and water bodies are excluded. Model performance is summarized at the 40th, 50th, 60th, 70th, 80th, and 90th percentiles of daily recall, computed over every day in the test period with at least one fire (103 days). The Q-th percentile is the daily recall value below which Q\% of the fire days fall. Equivalently, (100-Q)\% of the days performed at least that well. For instance, a 60th-percentile daily recall of 0.99 means 60\% of fire days had a daily recall $\leq$ 0.99, and the remaining 40\% had a daily recall $>$ 0.99. The results for each architecture are reported in Table \ref{tab:q_quantiles}.

\begin{table}
    \caption{Q-quantiles of daily recall (the fraction of fire pixels on a given day assigned fire probability > 0.5) for days with fire computed from full-map inference in fire seasons in 2020 and 2021 - a total of 103 days.}
    \centering
    \begin{tabular}{|ccccc|}
         \hline
          \begin{tabular}{@{}c@{}}\textbf{Q-Quantiles}  \\ \textbf{/Model} \end{tabular}  & \textbf{Basic CNN} & \textbf{Deeper CNN 1 }& \textbf{Deeper CNN 2} & \textbf{ConvLSTM} \\
         \hline
         \textbf{40\%} & 0.74 & 0.74 & 0.75 & 0.77 \\
         \hline
         \textbf{50\%} & 0.82 & 0.83 & 0.84 & 0.92 \\
         \hline
         \textbf{60\%} & 0.92 & 0.92 & 0.92 & 0.99 \\
         \hline
         \textbf{70\%} & 0.97 & 0.98 & 0.96 & 1.00 \\
         \hline
         \textbf{80\%} & 1.00 & 1.00 & 1.00 & 1.00 \\
         \hline
         \textbf{90\%} & 1.00 & 1.00 & 1.00 & 1.00 \\
         \hline
    \end{tabular}
    \label{tab:q_quantiles}
\end{table}

\begin{figure}
  \centering
  \includegraphics[width=\linewidth]{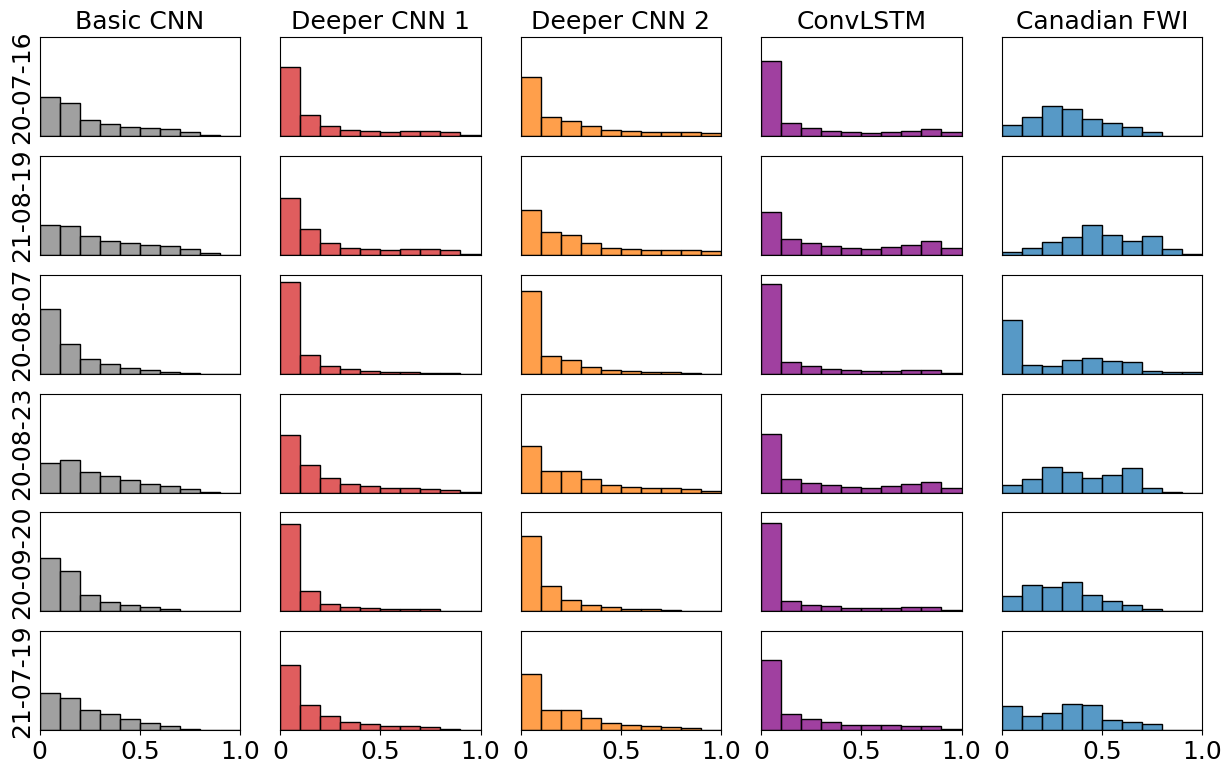}
  \caption{The distribution of the fire probability occurrence on no-fire days randomly selected. A positively skewed distribution indicates fewer false positives on that day compared to a negatively skewed distribution. ConvLSTM generally has a more positively skewed distribution than the CNN architectures. The right-most column shows the distribution of Canadian FWI on the same day, normalized by 70, which is classified as an extreme danger class.}
  \label{fig:fdi_dist_nofire}
\end{figure}

\begin{figure}
  \centering
  \includegraphics[width=\linewidth]{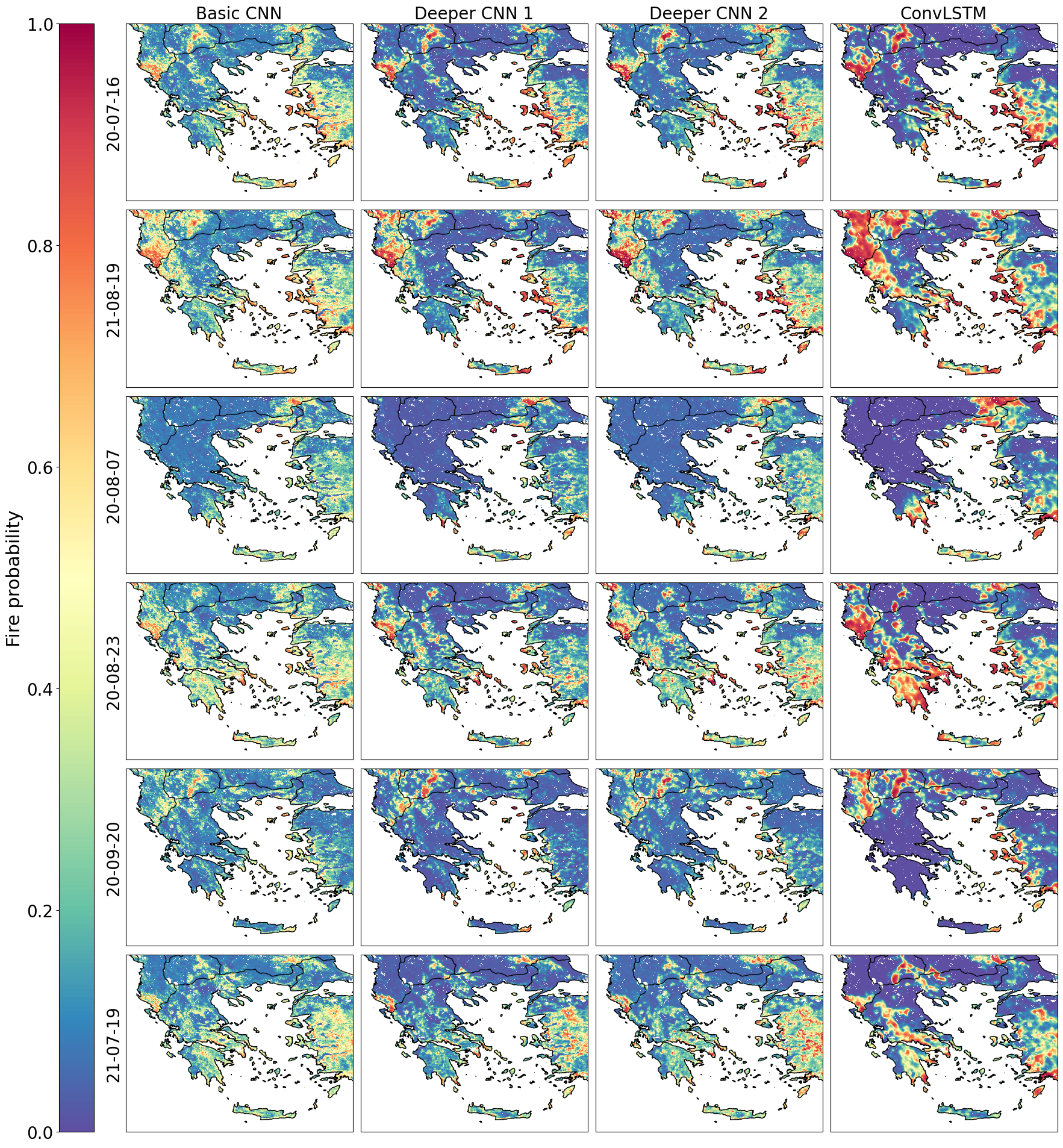}
  \caption{Region-wide inference for the no-fire days randomly selected (the same days as in Figure \ref{fig:fdi_dist_nofire}). Similarities between the maps indicate that each model is a noisy estimator of the underlying true fire probability; however, the ConvLSTM identifies fire-safe conditions far more confidently than the CNN variants (fire probabilities are far from the decision threshold of 0.5.}
  \label{fig:fdi_maps_nofire}
\end{figure}

 Fire activity maps on no-fire days are a useful sample for assessing false-alarm behavior: since no fires occurred, every high-probability location is unambiguously a false alarm. We compare models by the skewness of the fire-probability distribution over the entire map on these days. A distribution with its mass concentrated near zero and a thin tail extending toward higher probability values is positively skewed, indicating fewer false positives: most pixels are confidently assigned low fire probability, with only a small fraction extending into higher-probability territory. Conversely, a distribution with mass shifted toward higher probability values is negatively skewed and indicates more false positives, since a larger fraction of pixels are assigned elevated fire probability despite no fire having occurred. This analysis was carried out across all no-fire days in the test period, with consistent results throughout; for simplicity, we show six representative days. The distribution on the selected dates is shown in Figure \ref{fig:fdi_dist_nofire} along with the distribution of Canadian FWI \parencite{1987fwi} (as computed by \parencite{prapas2022firecube}) on the same day (normalized to the value of 70, which indicates extreme fire danger category) whereas the region-wide inference on the same dates from all four networks is shown in Figure \ref{fig:fdi_maps_nofire}. In the region-wide inferences shown in Figure \ref{fig:fdi_maps_nofire}, the visual similarities between the maps produced by different architectures are appreciable: different networks highlight similar regions at high fire danger.

\begin{figure}
  \centering
  \includegraphics[width=0.8\linewidth]{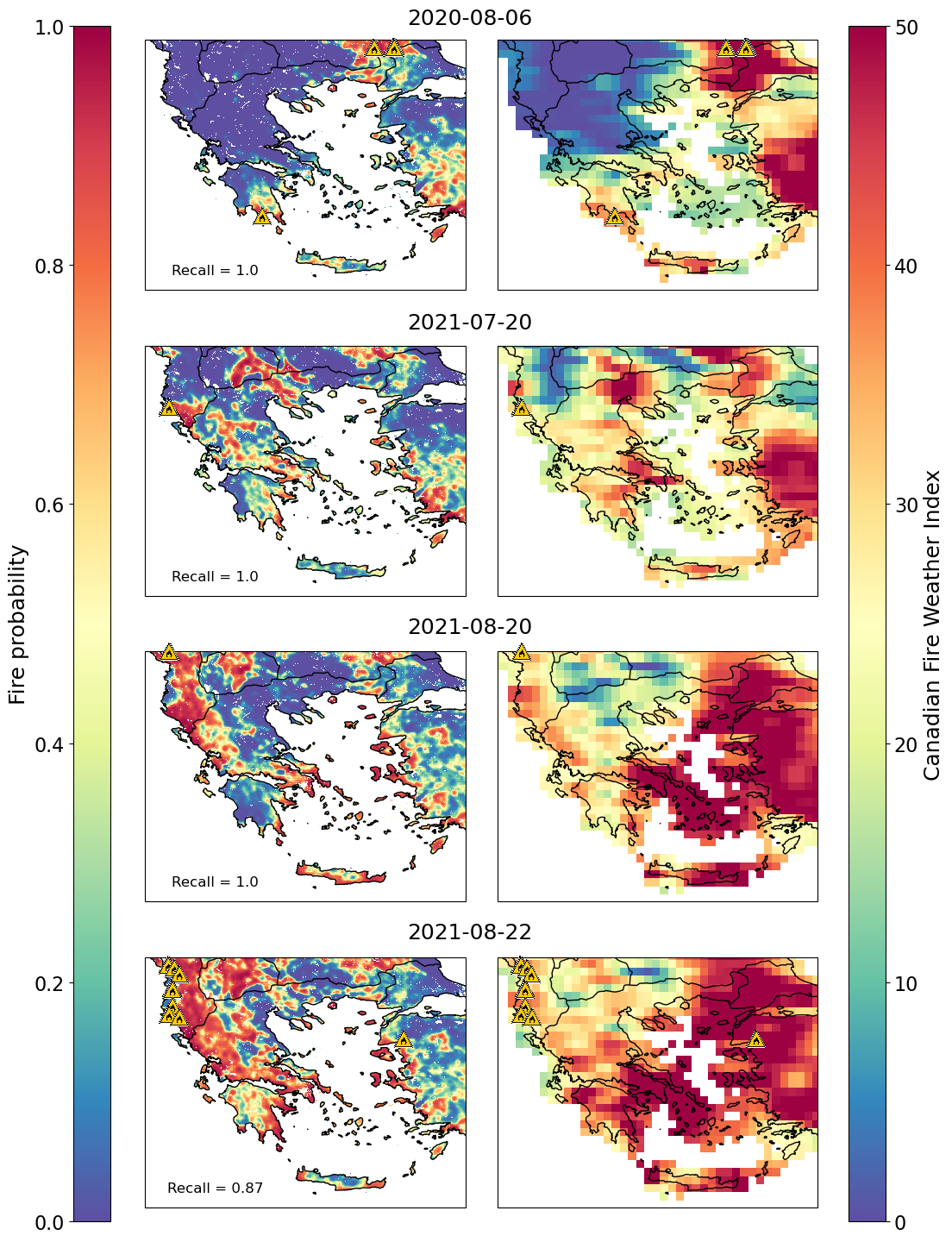}
  \caption{Comparison between fire probability maps from ConvLSTM (left column) and Canadian Fire Weather Index (right column), provided in the dataset \parencite{prapas2022firecube} dataset, on the four random days. In both plots, recorded fire events are represented by a danger icon.}
  \label{fig:fwi_vs_fdi}
\end{figure}

Figure \ref{fig:fwi_vs_fdi} compares ConvLSTM fire activity maps with the process-based Canadian Fire Weather Index (FWI; Wagner 1987) on four fire days. Consistent with prior findings, the data-driven maps are sharper and more spatially localized than FWI.

\subsection{Map-based analysis on ensemble-based approach}
\label{subsec:result_model_ens}
\begin{figure}[htbp]
  \centering
  \includegraphics[width=0.9\linewidth]{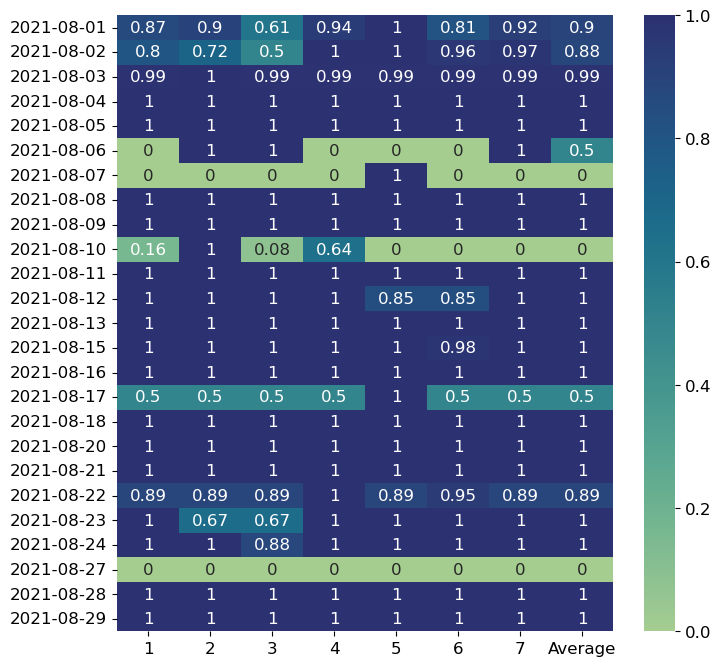}
  \caption{Daily recall computed on each day by each ensemble member is shown as a heatmap. The daily recall computed from the ensemble-averaged map is shown in the 8th column with the label ``Average''.}
  \label{fig:ens_recall}
\end{figure}

\begin{figure}[htbp]
  \centering
  \includegraphics[angle=90,width=1\linewidth]{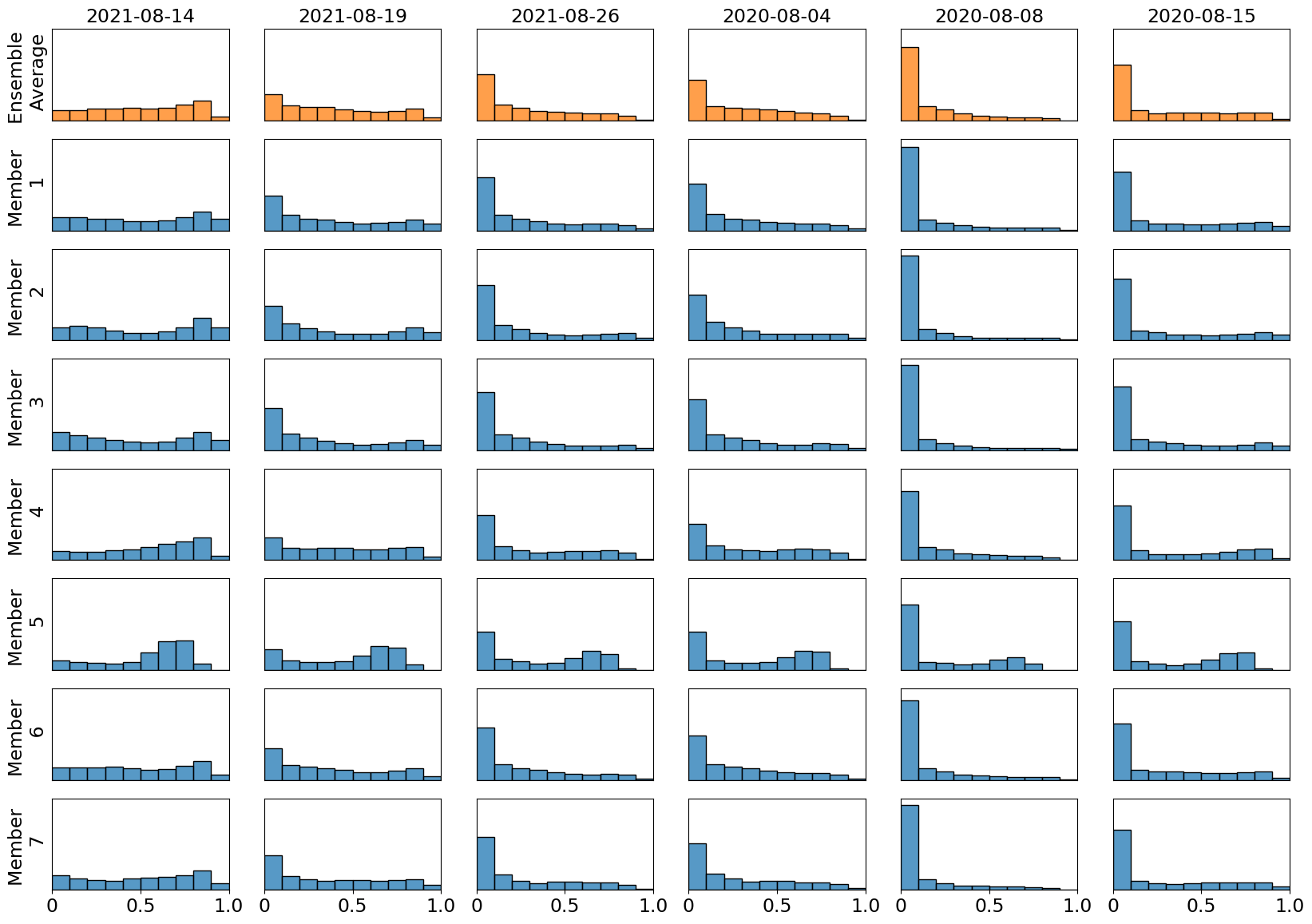}
  \caption{The probability distribution for six days. Since August 2021, there have been very few days with no fires; we also include some days in August 2020 in the analysis. The top plot (in orange) shows the distribution of the ensemble-averaged map.}
  \label{fig:ens_nofire}
\end{figure}

We evaluate the seven-member ConvLSTM ensemble (Section \ref{subsubsec:model_ensemble_training}) over August 2021, the most fire-active month in the dataset. We compute the fire-probability distribution on no-fire days for each ensemble member and for the ensemble average, following the same procedure as Section \ref{subsec:mapbased_eval}. As before, this analysis was carried out across the full no-fire days, with consistent results throughout. For demonstration, we show six representative days in August; since August 2021 contains too few no-fire days, three days are drawn from August 2020. The resulting distributions are shown in Figure \ref{fig:ens_nofire}.
We also report daily recall for each member and for the mean ensemble across the same period (Figure \ref{fig:ens_recall}).

\section{Discussion}\label{sec:discussion}
\subsection{Model performance: Test-dataset ranking vs. operational ranking}
Table \ref{tab:test_scores} and Table \ref{tab:q_quantiles} rank the four architectures in different order: On the curated test dataset, ConvLSTM ranks the lowest - Recall (0.87) and lowest Precision (0.73) - yet in the proposed framework, it is the strongest performer on both daily recall and false-alarm suppression (Figure \ref{fig:fdi_dist_nofire} and \ref{fig:fdi_maps_nofire}). From the percentiles in Table 3, three conclusions follow. First, all models detect more than 75\% of fires on 60\% of the days, indicating that CNN-based networks generally perform well at daily fire detection. Second, the ConvLSTM outperforms the CNN variants, achieving near-perfect detection (daily recall $>$ 0.99) on 40\% of fire days. Third, within the CNN family, depth does not improve daily recall, as indicated by nearly identical percentiles for Basic CNN, Deeper CNN 1, and Deeper CNN 2 in Table \ref{tab:q_quantiles}.

A high fire probability at a pixel where no fire was recorded does not by itself indicate model failure: the location may carry genuine latent risk that did not ignite that day. What is diagnostically relevant is the whole-map distribution. On a day when no ignition occurred anywhere in the region, a distribution with substantial mass at high probability reflects aggregated false-alarm load, among which not all are genuine risks on a day when no fire incidents occurred. No-fire days thus give a clean assessment of a model's false-alarm behaviour, and the skewness of the distribution over the entire map is its natural measure (Figure \ref{fig:fdi_dist_nofire}). Elevated false-alarm rates remain operationally costly, since they trigger unnecessary resource mobilization.

We refer to a map as noisy when many pixels are assigned probabilities near the 0.5 decision threshold, indicating the model is undecided between fire and no-fire. By this measure, Figure \ref{fig:fdi_dist_nofire} shows the ConvLSTM fire probability distribution concentrated near zero with a thin tail toward high probability (positive skew): most pixels sit far from the 0.5 threshold, so ConvLSTM maps are the least noisy of all architectures considered, consistent with Figure \ref{fig:fdi_maps_nofire}.

The same measure separates the CNN variants. The Basic CNN maps are noisier than those of Deeper CNN 1, while Deeper CNN 2 is slightly noisier than Deeper CNN 1; among the CNNs, Deeper CNN 1 is the least noisy, indicating it has the optimal depth for confidently separating fire from no-fire regions. Moreover, while the three CNN architectures perform comparably at identifying fire pixels (Table 3), confidently identifying no-fire regions requires capturing more subtle feature correlations, which likely explains why Deeper CNN 1 flags no-fire regions more confidently than the Basic CNN. This false-alarm analysis is difficult to obtain from test-dataset metrics alone, which is what makes the map-based framework an operationally novel method for model evaluation.

\subsection{Ensemble based analysis}
Let \(\Theta(t)\) denote the true, latent fire-probability field on day t, determined by the full (unobserved) state of the fire-relevant covariates. Each ConvLSTM ensemble member, trained from an independent random weight initialization, provides an estimator \(\hat{\Theta} \left(t\right)\) of this field. As Figure \ref{fig:ens_recall} shows, individual members' daily recall values differ substantially from one another on a given day, indicating that their underlying probability estimates are not identical. Ensemble-averaged fire probability reduces variance in architectural biases across members, causing the ensemble-averaged probability distribution on no-fire days to become more positively skewed than that of the noisiest individual members on all no-fire days in the test period. Figure \ref{fig:ens_nofire} shows this state of affairs for some selected days in August 2021. 

The same variance-reduction argument does not carry over to daily recall. Pixel-wise ensemble-averaged behavior, in effect, behaves as a majority vote: a fire detected confidently by only one or two members is outvoted by the rest and lost in the ensemble-averaged map, while a fire detected by a majority of members survives largely unchanged. We tested ensemble-median and ensemble-max aggregation as alternatives to ensemble-mean aggregation. Median aggregation improved neither daily recall nor the false-alarm distribution relative to the ensemble-average aggregation method. Max aggregation, by contrast, worsened the false-alarm distribution — an expected outcome, as it gives too much weight to the over-estimation of false alarm from a single ensemble member (only needs one member to flag a pixel with a false high fire probability).

The daily recall shortfall is not evenly distributed across the test period, as shown in Figure \ref{fig:ens_recall}. On 22 of the 25 evaluated days, a majority of the seven independently initialized members agree exactly on the daily recall value for that day — including, but not limited to, unanimous full detection (12 days) and unanimous complete failure (2021-08-27). This indicates the forecast is far from threshold on most days, regardless of whether the outcome is correct, as small variations in the model forecast would have caused the daily recall on the same day to fluctuate across ensemble members.

\section{Conclusions}\label{sec:conclusions}

This study evaluates CNN, its variants, and ConvLSTM architectures for daily forest fire forecasting over Greece, the Balkan peninsula, and western Turkey. The work proposes a novel approach for evaluation, focusing on operational performance, building on the work of \parencite{Kondylatos2022}. Standard machine-learning metrics are useful for monitoring training, but as argued in Section \ref{subsec:mapbased_eval}, test dataset class balance and potential feature-space overlap between fire and no-fire samples limit their relevance in operational deployment. Region-wide inference is therefore a more complete measure of a model's operational viability for daily fire forecasting. It also revealed a result the test-set metrics missed entirely: while standard metrics suggested all four networks performed comparably, with CNNs showing slightly higher Precision, region-wide inference-based analysis showed ConvLSTM achieving near-perfect detection (daily recall $>$ 0.99) on 40\% of fire days as it can exploit spatiotemporal correlations, integrating rapid meteorological change with slow-varying vegetation trends.
Systematic assessment of false positives flagged by the model is rarely explored in the literature. Our analysis goes beyond the standard ML paradigm in this aspect. ConvLSTM has a clear advantage over the model based on CNN, as it generally positively skews the probability distribution, indicating fewer false positives. The study further reveals that increasing complexity, from a Basic CNN to Deeper CNN 2, did not favor daily recall. However, Deeper CNN 1 was identified as having the optimal depth for confidently flagging no-fire regions. This suggests that simple layers might be sufficient for identifying a fire event, but capturing subtle correlations for ``safe'' conditions requires a more nuanced architecture.

We also examined a seven-member ConvLSTM ensemble. Averaging the members' probability maps consistently reduced false-alarm noise, producing a more confident and less noisy predictor of fire-free areas; the ensemble-averaged distribution showed fewer false positives than even the worst-performing individual member. The ensemble average did not improve daily recall, as the ensemble average behaved as an implicit majority vote, whereby a high daily recall in the ensemble average was only achieved if many members of the ensemble achieved a high daily recall.

Future work should establish region-wide inference as the primary evaluation method for daily fire forecasting, covering both detection and false-alarm behaviour. A natural next step is to formalize false-alarm assessment beyond distributional skewness towards a region-wide false-positive rate more quantitatively comparable across models.

\section{Funding and Acknowledgments}
This project has been partially funded by the ARCA (funded by Interreg IPAADRION programme under the European Regional Development Fund and IPA III, agreement number IPA-ADRION00107) and OfidiaPlus (funded by Interreg VI-A Greece-Italy 2021-2027 programme, agreement no. 600514).


\printbibliography





\end{document}